\documentclass[10pt,twocolumn,letterpaper]{article}

\usepackage{cvpr}              % To produce the CAMERA-READY version
\usepackage{graphicx}
\usepackage{amsmath}
\usepackage{amssymb}
\usepackage{booktabs}
\usepackage[table]{xcolor} % 放在导言区
\usepackage{algorithm}
\usepackage{algorithmic}

\usepackage{multirow}
\usepackage{makecell}
\usepackage{booktabs}
\usepackage[table]{xcolor}  % 支持表格底色
     
\definecolor{cvprblue}{rgb}{0.21,0.49,0.74}
\usepackage[pagebackref,breaklinks,colorlinks,allcolors=cvprblue]{hyperref}

\title{Degradation-Robust Fusion: An Efficient Degradation-Aware Diffusion Framework for Multimodal Image Fusion in Arbitrary Degradation Scenarios}

\author{
	Yu Shi\textsuperscript{1} \quad 
	Yu Liu\textsuperscript{1}\thanks{Corresponding author: Yu Liu.} \quad 
	Zhong-Cheng Wu\textsuperscript{1} \quad 
	Juan Cheng\textsuperscript{1} \quad
	Huafeng Li\textsuperscript{2} \quad
	Xun Chen\textsuperscript{3} \\
	\textsuperscript{1}Department of Biomedical Engineering, Hefei University of Technology \\ 
	\textsuperscript{2}Faculty of Information Engineering and Automation, Kunming University of Science and Technology \\
	\textsuperscript{3}School of Information Science and Technology, University of Science and Technology of China\\
	{\tt\small yushi@mail.hfut.edu.cn} \quad
	{\tt\small yuliu@hfut.edu.cn}
}
\begin{document}
\maketitle
\begin{abstract}
Complex degradations like noise, blur, and low resolution are typical challenges in real-world image fusion tasks, limiting the performance and practicality of existing methods. End-to-end neural network–based approaches are generally simple to design and highly efficient in inference, but their black-box nature leads to limited interpretability. Diffusion-based methods alleviate this to some extent by providing powerful generative priors and a more structured inference process. However, they are trained to learn a single-domain target distribution, whereas fusion lacks natural fused data and relies on modeling complementary information from multiple sources, making diffusion hard to apply directly in practice. To address these challenges, this paper proposes an efficient degradation-aware diffusion framework for image fusion under arbitrary degradation scenarios. Specifically, instead of explicitly predicting noise as in conventional diffusion models, our method performs implicit denoising by directly regressing the fused image, enabling flexible adaptation to diverse fusion tasks under complex degradations with limited steps. Moreover, we design a joint observation model correction mechanism that simultaneously imposes degradation and fusion constraints during sampling to ensure high reconstruction accuracy. Experiments on diverse fusion tasks and degradation configurations demonstrate the superiority of the proposed method under complex degradation scenarios. Code: \url{https://github.com/YShi-cool/DRFusion}.
\end{abstract}    
\section{Introduction}
\label{sec:intro}
\begin{figure}[htbp]
	\centering
	\includegraphics[width=1\linewidth]{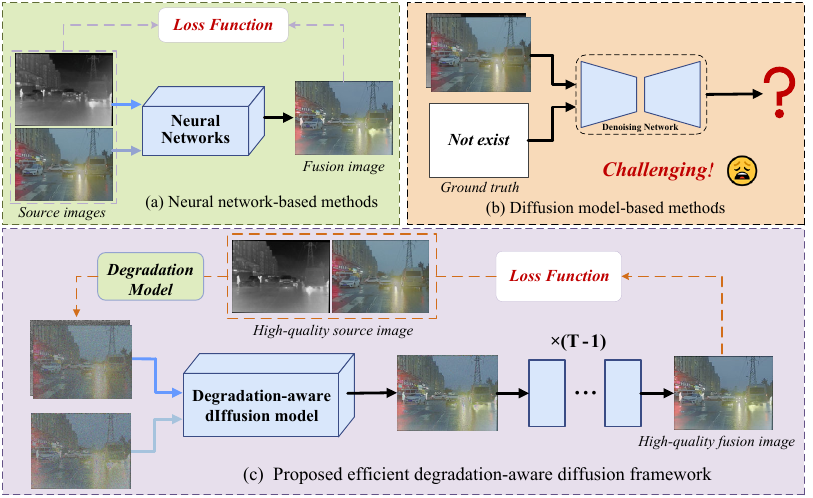} % 替换为你的图片文件名
	\caption{Comparison of fusion strategies under different degradation scenarios: (a) methods based on neural networks; (b) existing diffusion-based methods; (c) the proposed degradation-aware diffusion framework from this work.}
	\label{fig1}
\end{figure}

With the rapid development of multimodal imaging technologies in fields such as medical imaging, industrial vision, and video surveillance, image fusion techniques have become critical for enhancing image quality and improving visual understanding. 
However, most existing fusion methods assume that source images are of high quality, which overlooks the prevalent degradation phenomena in real-world imaging environments, such as noise, motion blur, and insufficient resolution. These degradation factors introduce distortions and result in the loss of critical information, significantly limiting both the accuracy of the fusion results and the practical robustness of the system.

The traditional “restoration + fusion” paradigm makes the fusion result highly dependent on restoration quality, while its decoupled design can introduce error accumulation across stages and complicate deployment. These issues motivate designing restoration and fusion within a unified framework \cite{yin2013simultaneous, li2021joint, li2018joint}. As illustrated in Fig. \ref{fig1} (a), end-to-end neural network–based fusion methods construct a parametric mapping from multi-source inputs to a fusion output, which is optimized using loss functions that jointly constrain restoration and fusion quality \cite{li2021different, huang2024dednet, chen2024mdbfusion}. These methods are usually simple to design and computationally efficient. However, their black-box nature limits interpretability, and the achievable reconstruction accuracy is highly dependent on the choice and design of the loss functions.

Diffusion models \cite{ho2020denoising, he2025diffusion, croitoru2023diffusion}, a class of generative models, have demonstrated exceptional performance in image generation and reconstruction tasks due to their superior distribution modeling capabilities and stability. Unlike the black-box mapping of deep neural networks, diffusion models offer superior interpretability and higher fusion accuracy through their solid theoretical foundations. This iterative mechanism is highly beneficial to image fusion, making the progressive aggregation of multimodal information transparent. Additionally, such step-wise refinement inherently guarantees the stability of the fusion process. However, the following inherent limitations of diffusion models make it difficult to apply them to image fusion under degraded conditions. First, diffusion models are trained to learn a target data distribution, whereas image fusion lacks naturally available fused data to support such training \cite{yi2024diff, wang2024uud}, as shown in Fig. 2 (b). Second, standard diffusion models operate on a single-domain distribution, whereas fusion requires modeling complementary information from multiple sources, which calls for an additional formulation that explicitly links cross-modal information, fusion objectives, and the probabilistic model \cite{zhao2023ddfm}. In addition, the computational cost of iterative sampling cannot be ignored. Existing methods either address only specific degradations \cite{xu2024simultaneous} or rely on independently pre-trained restoration models \cite{tang2024drmf}, yet they still do not provide an effective and flexible framework capable of handling diverse fusion tasks under complex degradation scenarios.

To address these challenges, this paper proposes a degradation-aware diffusion framework that unifies restoration and fusion for multimodal image fusion under arbitrary degradations. In contrast to the explicit noise-prediction paradigm in standard diffusion models, we adopt a direct fused-image regression formulation that implicitly encodes the denoising process, enabling the model to flexibly handle diverse fusion tasks under complex degradation conditions like the conventional end-to-end frameworks. By removing explicit noise prediction, the framework can achieve competitive results within a limited number of diffusion steps, leading to a substantial improvement in inference efficiency. Moreover, we design a joint observation correction mechanism that injects both degradation and fusion constraints into the diffusion sampling process, forcing intermediate samples to remain aligned with the degradation model while preserving complementary cross-modal information, thereby ensuring high reconstruction accuracy of the fusion results.
Our contributions can be summarized as follows:
\begin{itemize}
		\item We propose an efficient degradation-aware diffusion framework that jointly models degradation and multimodal image fusion within a single process. By directly regressing the fused image instead of predicting noise, our approach flexibly adapts to diverse tasks and achieves competitive performance in few diffusion steps.
		\item We design a joint observation correction mechanism that injects both degradation and fusion constraints into the diffusion sampling process, enforcing consistency with the degradation model while preserving complementary cross-modal information to improve the reconstruction accuracy of the fusion results.
		\item We conduct extensive experiments across diverse fusion tasks and complex degradation scenarios, demonstrating superior quantitative and visual performance.
\end{itemize}
\section{Related Work}
%-------------------------------------------------------------------------
\subsection{Multimodal Image Fusion}
Based on their decomposition approaches, early image fusion methods primarily rely on multi-scale transforms \cite{liu2015general, li2021infrared} and sparse representation \cite{liu2015simultaneous, liu2016image, li2024infrared}. In recent years, deep learning (DL)-based methods \cite{liu2017multi, yi2024text, zhou2024general}, have gradually replaced traditional methods to become the mainstream in research. With the evolution of DL, these methods have progressed from convolutional neural network (CNN)-based approaches \cite{amin2019ensemble, xu2021emfusion, liu2024coconet} to attention-based architectures, including Transformer \cite{liu2024mm, zhao2024equivariant, zhao2023cddfuse} and the more recent Mamba models \cite{xie2024fusionmamba, cao2024novel, zhu2025mamba}. Building upon these architectural advances, recent studies have shifted focus toward task-oriented enhancements. For example, to improve the practicality and ease of deployment of fusion models, several unified frameworks have been proposed \cite{liu2023lightweight, wang2021unfusion}. In addition, some approaches integrate downstream tasks into the fusion process to enhance the overall applicability and effectiveness of the method \cite{liu2022glioma,liu2024task, yang2025instruction}. To address common degradation issues that may occur during the imaging process, such as noise, low resolution, and imprecise alignment, some studies have integrated degradation modeling and image fusion into a single framework \cite{li2021joint, ma2020ddcgan, 10856402}. 

\begin{figure*}[htbp]
	\centering
	\includegraphics[width=0.9\textwidth]{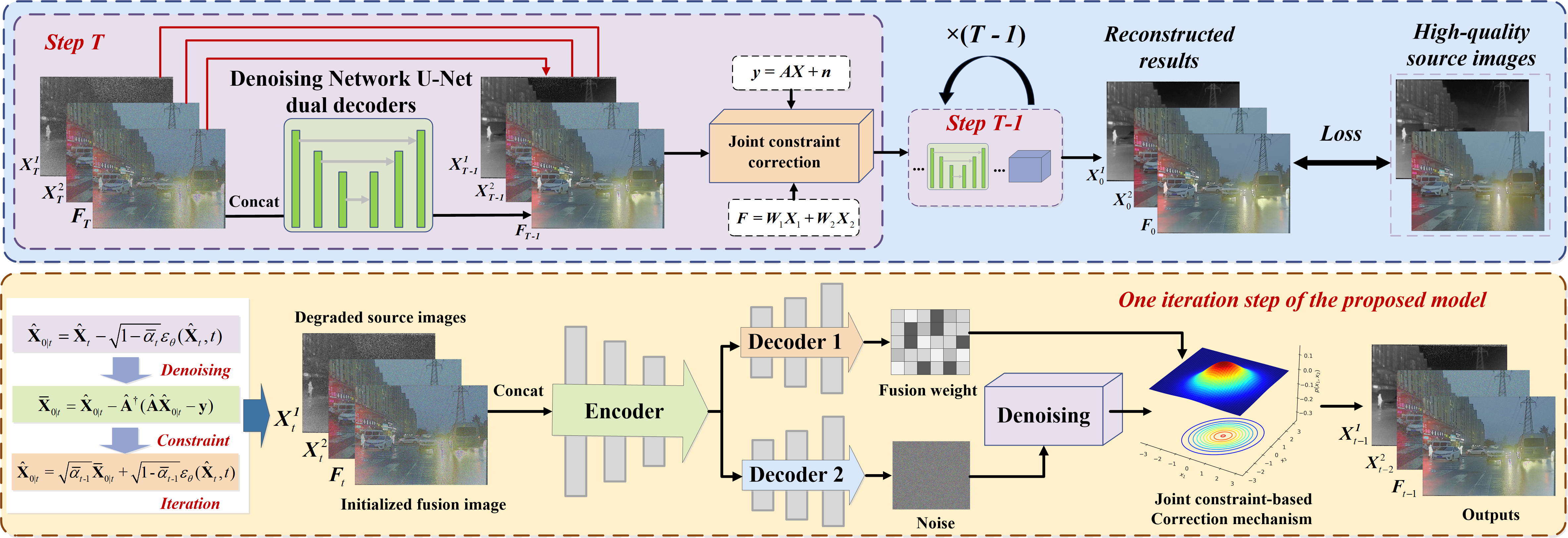} % 替换为你的图片文件名
	\caption{The proposed framework for multimodal image fusion under various degradation scenarios in this work.}
	\label{fig2}
\end{figure*}

Due to their strong generative capabilities, robust generalization, and structured inference, diffusion models are increasingly popular in image fusion \cite{yang2025lfdt, yi2024diff, pan2024dif}. While directly applicable to supervised tasks \cite{li2024fusiondiff}, diffusion models struggle in practical, unsupervised fusion scenarios due to their reliance on target data distributions. To address this challenge, existing methods have proposed several strategies, such as incorporating task-specific mathematical formulations into the diffusion framework \cite{zhao2023ddfm, shi2024vdmufusion, xu2025flexid}, designing dedicated decoders to reconstruct fusion results from intermediate features \cite{yue2023dif}, or generating pseudo labels to guide the training \cite{xu2024simultaneous, 11050979}. Despite recent advances, current diffusion-based fusion methods remain constrained by standard frameworks, limiting their effectiveness in complex scenarios like joint restoration and fusion.

%-------------------------------------------------------------------------
\subsection{Restoration Oriented Diffusion Models}

Diffusion models have emerged as a powerful paradigm for image restoration tasks, including denoising \cite{wang2023dr2}, deblurring \cite{xia2023diffir}, and super‑resolution \cite{wang2025rap}.Compared to existing algorithms \cite{li2024contourlet, li2025difiisr}, their iterative denoising and strong generative priors offer greater flexibility in modeling complex data distributions. Fei et al. proposed generative diffusion prior \cite{fei2023generative}, demonstrating how a pre-trained diffusion model can serve as a unified unsupervised prior for multiple restoration and enhancement tasks. In the context of super-resolution, SinSR \cite{wang2024sinsr} further shows that diffusion-based methods can approach or surpass state-of-the-art performance while requiring fewer inference steps. For deblurring, DDNM \cite{wang2022zero} incorporates a pseudoinverse-based consistency step that embeds the blur model into each denoising iteration, keeping samples aligned with the degraded input. However, unlike single-image restoration, fusion tasks must integrate complementary information from multiple degraded sources, necessitating more elaborate designs beyond standard pipelines.

\section{Method}

\subsection{Fusion-Oriented Diffusion Framework}

In standard diffusion models as shown in Supplementary Material A, a noise prediction network is first pre-trained so that it can estimate the noise injected at any diffusion timestep $t$. The noise added at timestep $t$ and the corresponding noisy image can typically be obtained as follows:
\begin{equation}
	p(\bf{x}_t|{\bf{x}_0}) = \mathcal{N}({\bf{x}_t};\sqrt {{{\bar \alpha }_t}}  \cdot {\bf{x}_0},(1 - {\bar \alpha _t}) \cdot {\bf{I}}),
\end{equation}
where ${\alpha _t} = 1 - {\beta _t}$ and ${\bar \alpha _t} = \prod\nolimits_{i = 1}^t {{\alpha _t}} $, ${\beta _1},{\beta _2},...,{\beta _T} \in [0,1)$ denote a set of hyperparameters that control the variance schedule over diffusion timesteps. To facilitate training, using the reparameterization trick, the above equation can be rewritten as:
\begin{equation}
	p({\bf{x}_t}|{\bf{x}_0}) = \mathcal{N}({\bf{x}_t};\sqrt {{{\bar \alpha }_t}}  \cdot {x_0},(1 - {\bar \alpha _t}) \cdot {\bf{I}}).
\end{equation}
To achieve accurate noise estimation, the network typically needs to be trained for a large number of iterations, and the diffusion period $T$ is usually set to be relatively large, which further increases the overall computational cost. This architectural design requires access to the target data distribution during training, effectively restricting diffusion models to supervised settings where clean target samples are available. Moreover, they are typically trained to model a single-domain distribution, making it nontrivial to directly extend them to multi-input tasks such as multimodal image fusion.

To address the above issues, inspired by \cite{chen2025invertible}, we propose a new diffusion framework specifically designed for image fusion, as illustrated in Fig.~\ref{fig2}. We discard the explicit noise-prediction step in standard diffusion models and retain only the reverse process, directly mapping the inputs to the fusion output through a limited number of diffusion iterations.
\begin{equation}
{F_\theta } = f_\theta ^T \to f_\theta ^{T - 1} \to  \cdots f_\theta ^0,
\end{equation}
where $F_\theta$ denotes the proposed diffusion framework, $f_\theta ^t$ denotes the diffusion iteration at timestep $t$. At each diffusion step $f_\theta ^{t}$, an accelerated DDIM-based sampling scheme is employed:
\begin{equation}
{{\hat{\bf{x}}}_{0 | t}} = {{{\bf{\hat x}}}_t} - \sqrt {1 - {{\bar \alpha }_t}} {\varepsilon _\theta }({{\bf{\hat x}}_t},t),
\end{equation}
\begin{equation}
	{{\bf{\hat x}}_{t - 1}} = \sqrt {{{\bar \alpha }_{t - 1}}} {{\bf{\hat x}}_{0|t}} + \sqrt {1 - {{\bar \alpha }_{t - 1}}} {\varepsilon _\theta }({{\bf{\hat x}}_t},t).
\end{equation}
This design makes the diffusion model closer in architecture to an end-to-end neural network architecture and brings several advantages. First, the direct input-to-output mapping allows the proposed method to handle fusion in a self-supervised manner, alleviating the lack of fusion labels. Second, multi-input fusion tasks can be jointly optimized within a single framework by imposing appropriate loss constraints on the fusion output with respect to each source input. Third, since the model no longer predicts noise explicitly but hides it in intermediate representations, it can be combined with accelerated samplers such as DDIM to obtain high-quality results within a limited number of diffusion steps, thereby improving inference efficiency.

\subsection{Joint Observation Correction Mechanism}
Although diffusion models are capable of producing high-quality results, their sampling process is inherently stochastic. This randomness is most pronounced when using standard stochastic samplers like DDPM. In contrast, accelerated schemes such as DDIM can yield nearly deterministic sampling trajectories. Nevertheless, it's crucial to remember that the underlying model is still learning a data distribution, not a single deterministic mapping. Therefore, some works introduce additional constraints into the DDIM sampling process to guide the generation towards solutions that are better aligned with the desired task objectives \cite{wang2022zero, chen2025invertible}. Motivated by these constraint-based strategies but aiming for a more unified and flexible solution, we propose a degradation-aware diffusion framework tailored for multimodal image fusion under complex degradations.

\textbf{Degradation-Aware Diffusion Correction.} The classical image degradation model can be formulated as follows:
\begin{equation}
{\bf{y}} = {\bf{AX}} + {\bf{n}},
\end{equation}
where ${\bf{y}}\in{\mathbb{R}^{d \times 1}}$ denotes the observed image, $\bf{A}\in{\mathbb{R}^{d \times D}}$ is the degradation matrix (operator), $\bf{X}\in{\mathbb{R}^{D \times 1}}$ represents the underlying clean image, and $\bf{n}\in{\mathbb{R}^{d \times 1}}$ is the noise term. We temporarily ignore the noise term. In the previous diffusion sampling step, we obtain a denoised image estimate ${\hat{\bf {x}}}_{0|t}$, which, however, does not necessarily satisfy the above observation constraint ${\bf{y}} = {\bf{AX}}$. Therefore, starting from the estimate obtained in the previous diffusion step, we aim to find a new solution that satisfies the above constraint, which requires performing a projection onto the constraint set:
\begin{equation}
{{\bf{x}}^ \star } = \arg \mathop {\min }\limits_{\bf{z}} ||{\bf{z}} - {{\bf{x}}_{0|t}}|{|^2}\;\;{\rm{s}}{\rm{.t}}{\rm{.}}\;{\bf{Az}} = {\bf{y}}.
\end{equation}
Geometrically, this is equivalent to projecting the point ${\bf{x}}_{0|t}$ onto the subspace defined by ${\bf{Az}} = {\bf{y}}$. The solution to this problem is:
\begin{equation}
\bf{x}^\star=\bf{x}-A^\dagger(A\bf{x}-y),
\end{equation}
where $\bf{A}^\dagger$ is the Moore–Penrose pseudoinverse, $r=\bf{Ax} - \bf{y}$ is defined as the residual and $\delta  = {\bf{A^\dagger}}r$ as the correction term. Therefore, the projection step is given by subtracting the correction from the basic solution $\bf{x}^\star=\bf{x}_{0|t}-\delta$. For traditional restoration tasks, the above formulation is straightforward to solve, since every term is known. However, for the image fusion problem under complex degradation conditions considered in this work, the situation becomes much more complicated. Existing methods mainly address single-image degradation, whereas fusion involves multiple inputs. In addition, the fusion image does not have a corresponding degraded observation. These factors together pose significant challenges for our solution (see Supplementary Material B for detailed analysis).

\textbf{Joint Observation Model.} This paper proposes a joint observation model to address the above problems. First, we rewrite the multiple inputs in the form of a joint variable $\left[ {{{\bf{X}}_1},{{\bf{X}}_2},{{\bf{X}}_f}} \right]$, and the degradation constraints of the two source images together with the fusion constraint can be expressed as follows:
\begin{equation}
	{\bf{y}_1} = {\bf{A}}_1{\bf{X}}_1,
\end{equation}
\begin{equation}
	{\bf{y}_2} = {\bf{A}}_2{\bf{X}_2},
\end{equation}
 \begin{equation}
 	{{\bf{X}}_f} = {{\bf{W}}_1} * {{\bf{X}}_1} + {{\bf{W}}_2} * {{\bf{X}}_2}.
 \end{equation}
 By moving ${{\bf{X}}_f}$ in Eq. (11) to the right-hand side, the original position of the fused image is replaced by a zero matrix, which indicates that we do not need to obtain the fused image in advance. The above model can be reformulated in the following joint matrix form:
\begin{equation} 
 \left[ \begin{array}{l}
 	{{\bf{y}}_1}\\
 	{{\bf{y}}_2}\\
 	{{\;\bf{0}}}
 \end{array} \right] = \left[ \begin{array}{l}
 	\;{{\bf{A}}_1}\;\;\;\;\;\;\;\;\;0\;\;\;\;\;\;\;\;\;0\\
 	\;\;0\;\;\;\;\;\;\;\;\;\;{{\bf{A}}_2}\;\;\;\;\;\;\;0\\
 	- {{\bf{W}}_1}\;\; - {{\bf{W}}_2}\;\;\;\;\;{\bf{I}}
 \end{array} \right]\left[ \begin{array}{l}
 	{{\bf{X}}_1}\\
 	{{\bf{X}}_2}\\
 	{{\bf{X}}_f}
 \end{array} \right],
 \end{equation}
 where ${{\bf{A}}_1}$ and ${{\bf{A}}_2}$ denote the degradation operators associated with the two source images, ${{\bf{W}}_1}$ and ${{\bf{W}}_2}$ are the fusion operators, and $\bf{I}$ is the identity matrix. The above joint observation model not only overcomes the limitation of previous formulations that can only be applied to single-image restoration but also eliminates the need to explicitly obtain the fusion observation, further enabling the simultaneous execution of both restoration and fusion under a unified framework. The resulting joint degradation matrix takes the form of a block matrix:
 \begin{equation} 
 	{\bf{\hat A}} = \left[ \begin{array}{l}
 		\;{{\bf{A}}_1}\;\;\;\;\;\;\;\;\;0\;\;\;\;\;\;\;\;\;0\\
 		\;\;0\;\;\;\;\;\;\;\;\;\;{{\bf{A}}_2}\;\;\;\;\;\;\;0\\
 		- {{\bf{W}}_1}\;\; - {{\bf{W}}_2}\;\;\;\;\;{\bf{I}}
 	\end{array} \right].
 \end{equation}
 However, when applying it to the constrained procedure in Eq. (8), one issue remains: the pseudoinverse of the joint degradation matrix is difficult to compute. Although there are many ways to obtain a pseudoinverse, explicitly computing it is clearly impractical, as it would incur prohibitive computational cost and memory usage. Therefore, we propose an equation-based approach to compute the pseudoinverse implicitly by solving the corresponding linear system. By solving Eqs. (9)–(11) separately and then arranging the resulting solutions into a joint matrix in the same form as Eq. (12), the following formula can be obtained:
 \begin{equation} 
 	\left[ \begin{array}{l}
 		{{\bf{X}}_1}\\
 		{{\bf{X}}_2}\\
 		{{\bf{X}}_f}
 	\end{array} \right] = \left[ \begin{array}{l}
 		\;\;\;{{\bf{A}}_1}^\dagger\;\;\;\;\;\;\;\;0\;\;\;\;\;\;\;\;\;\;0\\
 		\;\;\;\;0\;\;\;\;\;\;\;\;\;\;{{\bf{A}}_2}^\dagger\;\;\;\;\;\;\;0\\
 		{{\bf{W}}_1}{{\bf{A}}_1}^\dagger\;\;  {{\bf{W}}_2}{{\bf{A}}_2}^\dagger\;\;\;{\bf{I}}
 	\end{array} \right]\left[ \begin{array}{l}
 	{{\bf{y}}_1}\\
 	{{\bf{y}}_2}\\
 	{{\;\bf{0}}}
 	\end{array} \right].
 \end{equation}
 In this way, we obtain a pseudoinverse of the joint degradation matrix ${\bf{\hat A}}$ that satisfies the Moore–Penrose conditions:
 \begin{equation} 
 	{\bf{\hat A}}^\dagger = \left[ \begin{array}{l}
 		\;\;\;{{\bf{A}}_1}^\dagger\;\;\;\;\;\;\;\;0\;\;\;\;\;\;\;\;\;\;0\\
 		\;\;\;\;0\;\;\;\;\;\;\;\;\;\;{{\bf{A}}_2}^\dagger\;\;\;\;\;\;\;0\\
 		{{\bf{W}}_1}{{\bf{A}}_1}^\dagger\;\;  {{\bf{W}}_2}{{\bf{A}}_2}^\dagger\;\;\;{\bf{I}}
 	\end{array} \right].
 \end{equation}
 By substituting Eq. (15) into Eq. (8), we obtain the final joint observation constraint, which is then embedded into the DDIM framework in Eqs. (4)–(5) to derive the final diffusion iteration form:
 \begin{equation}
 	{{\hat{\bf{x}}}_{0|t}} = {{{\bf{\hat x}}}_t} - \sqrt {1 - {{\bar \alpha }_t}} {\varepsilon _\theta }({{\bf{\hat x}}_t},t),
 \end{equation}
 \begin{equation}
 	{{\bar{\bf{x}}}_{0|t}}={\bf{\hat x}}_{0|t}-{\bf{\hat A}}^\dagger(\bf{\hat A}{\bf{\hat x}}_{0|t}-y),
 \end{equation}
 \begin{equation}
 	{{\bf{\hat x}}_{t - 1}} = \sqrt {{{\bar \alpha }_{t - 1}}} {{\bf{{\bf{\bar x}}}}_{0|t}} + \sqrt {1 - {{\bar \alpha }_{t - 1}}} {\varepsilon _\theta }({{\bf{\hat x}}_t},t).
 \end{equation}
 By sequentially performing the above three stages, we obtain a degradation-aware diffusion framework that is specifically designed for image fusion. It is worth noting that, to achieve better fusion performance, we do not adopt fixed fusion weights, instead, we learn them in a data-driven manner. Specifically, as illustrated in Fig. \ref{fig2}, the noise predictor is designed as a multi-task architecture that, in addition to predicting the noise, also outputs a weight map $\bf{W}_1$. We then enforce $\bf{W}_1 + \bf{W}_2=1$ to obtain the two complementary fusion weights. The overall workflow of our proposed method is summarized in Algorithm 1 of the supplementary material.
 
 \textbf{Managing More Complex Degradation Scenarios.} The above discussion is based on the noise-free case. For the noisy setting, we only need to modify Eq. (17) as follows:
 \begin{equation}
 	{{\bf{\bar{x}}}_{0|t}}={\bf{\hat x}}_{0|t}-{{\bf\Sigma} _t}{{\bf\hat A}}^\dagger({\bf{\hat A}}{\bf{\hat x}}_{0|t}-\bf{y}),
 \end{equation}
 where ${{\bf\Sigma} _t} \in {\mathbb{R}^{D \times D}}$ is used to scale the correction term in order to reduce the influence of noise \cite{wang2022zero}. Moreover, for more complex degradation scenarios, such as combinations of noise, blur, and low resolution simultaneously present in both source images, the proposed joint observation model can still handle them effectively. For instance, if source image is subjected to several types of composite degradation ${{\bf{A}}} = {\bf{A}}_1{\bf{A}}_2 \cdots {\bf{A}}_n$, its corresponding pseudoinverse operator can be written accordingly ${{\bf{A}}}^\dagger = {\bf{A}}_n^{\dagger}{\bf{A}}_{n - 1}^{\dagger} \cdots {\bf{A}}_1^{\dagger}$ and directly substituted into Eq. (15) to obtain a valid solution, which is both concise and efficient (see Supplementary Material C for detailed analysis).
 
 \subsection{Loss Function}
 Unlike previous diffusion based fusion methods, the proposed framework no longer explicitly predicts the noise but directly regresses the reconstructed image. This design allows us to adopt commonly used unsupervised losses in the image fusion literature to accomplish the above task. Specifically, the overall loss function consists of two components: a reconstruction loss on the source images and a fusion loss:
  \begin{equation}
 	{L_{total}} = {L_{rec}} + \lambda {L_f},
 \end{equation}
 where $ {L_{rec}}$ is defined as:
   \begin{equation}
 	{L_{rec}} = ||{{\bf{X}}_1} - {{\bf{\bar X}}_1}||_1 + ||{{\bf{X}}_2} - {{\bf{\bar X}}_2}||_1,
 \end{equation}
 where ${{\bf{X}}_1}$ and ${{\bf{X}}_2}$ represent the two reconstructed source images, ${{\bf{\bar X}}_1}$ and ${{\bf{\bar X}}_2}$ denote the labels of the two source images, and $||\cdot||_1$ denotes the $L1$ norm. For different fusion tasks, we adopt different fusion losses. In the case of infrared–visible image fusion, the loss is defined as follows:
 \begin{equation}
 	{L_{rec}} = ||{{\bf{X}}_1} - {{\bf{\bar X}}_1}||_1 + ||{{\bf{X}}_2} - {{\bf{\bar X}}_2}||_1.
 \end{equation}
 For different fusion tasks, we adopt different fusion losses. In the case of infrared–visible image fusion, the loss is defined as follows:
 \begin{equation}
 	\begin{split}
 		L_f &= ||\mathbf{X}_f - \max (\mathbf{\bar X}_1, \mathbf{\bar X}_2)||_1 \\
 		&\quad + \gamma ||\nabla \mathbf{X}_f - \max (\nabla \mathbf{\bar X}_1, \nabla \mathbf{\bar X}_2)||_1,
 	\end{split}
 \end{equation}
 where $\nabla$ denotes the gradient calculation operation, $\gamma$ is the hyperparameter trade-off. For the medical image fusion problem, the fusion loss is defined as follows:
 \begin{equation}
 	{L_f} = \sum\limits_{i = 1}^2 {||{{\bf{X}}_f} - {{{\bf{\bar X}}}_i}|{|_1} + \phi (1 - \mathrm{SSIM}({{\bf{X}}_f},{{{\bf{\bar X}}}_i})} ),
 \end{equation}
 where $\mathrm{SSIM}(\cdot)$ represents structural similarity calculation, $\phi$ is a hyperparameter. The proposed framework flexibly combines different loss terms according to the fusion task at hand, enabling task-oriented optimization instead of being tied to a single noise-prediction objective. This breaks the rigidity of traditional diffusion-based fusion frameworks.
%--------------------------------------------------------------------------
\section{Experiments}

\begin{figure*}[htbp]
	\centering
	\includegraphics[width=0.9\textwidth]{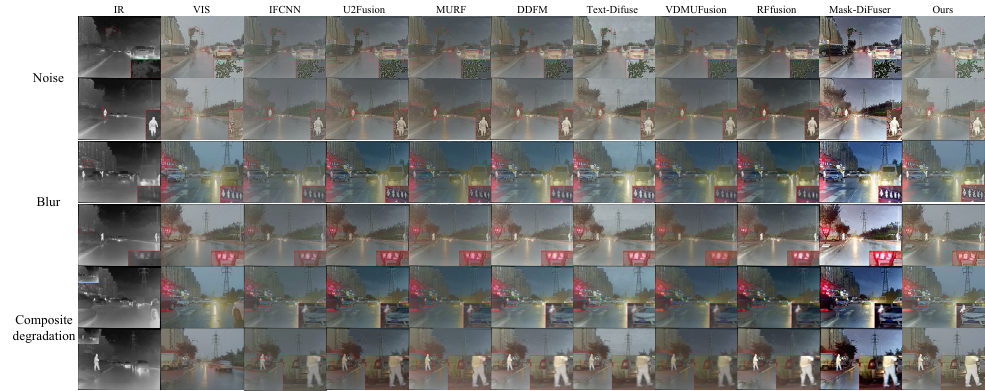} % 替换为你的图片文件名
	\caption{Qualitative results of different fusion methods on M3FD dataset. For the comparison methods, we first use corresponding restoration algorithms (e.g., denoising, deblurring, super-resolution) to restore the images, and then apply the respective fusion methods. The proposed method directly reconstructs the fusion results from the degraded source images. In the third degradation scenario, the infrared image is of low resolution and we have enlarged it, the original image is shown in the top-left corner.}
	\label{fig3}
\end{figure*}

\renewcommand{\arraystretch}{1.5}
\begin{table*}[t]
	\centering
	\caption{Objective fusion metrics of various methods under different degradation scenarios on the M3FD dataset (Bold and gray background: best result in each column; underline: second best).}
	\scalebox{0.56}{
		\begin{tabular}{c|cccccc|cccccc|cccccc}
			\hline
			\multirow{2}{*}{Methods} 
			& \multicolumn{6}{c}{Noise} 
			& \multicolumn{6}{c}{Blur} 
			& \multicolumn{6}{c}{Composite degradation} \\
			\cmidrule(r){2-7}\cmidrule(r){8-13}\cmidrule(r){14-19}
			& $Q_{MI}$ & $Q_{NCIE}$ & $Q^{AB/F}$ & $Q_{P}$ & $Q_{CB}$ & $Q_{W}$
			& $Q_{MI}$ & $Q_{NCIE}$ & $Q^{AB/F}$ & $Q_{P}$ & $Q_{CB}$ & $Q_{W}$
			& $Q_{MI}$ & $Q_{NCIE}$ & $Q^{AB/F}$ & $Q_{P}$ & $Q_{CB}$ & $Q_{W}$ \\
			\hline
			IFCNN
			& 0.2205 & 0.8036 & 0.4056 & {0.2115} & 0.4015 & \underline{0.7778}
			& 0.3378 & 0.8051 & 0.1042 & \underline{0.1057} & 0.3491 & 0.4969
			& 0.3176 & 0.8048 & 0.0965 & 0.0571 & 0.3635 & 0.4708 \\
			
			U2Fusion
			& 0.2454 & 0.8039 & \cellcolor{gray!20}\textbf{0.4281} & 0.1817 & \cellcolor{gray!20}\textbf{0.4804} & 0.7272
			& 0.2891 & 0.8045 & 0.1389 & 0.0696 & \underline{0.4205} & \underline{0.5035}
			& 0.2817 & 0.8044 & 0.1215 & 0.0491 & \cellcolor{gray!20}\textbf{0.4401} & 0.5507 \\
			
			MURF
			& 0.2270 & 0.8038 & 0.2630 & 0.0975 & 0.4383 & 0.5907
			& 0.3335 & 0.8054 & 0.1413 & 0.0793 & 0.3966 & 0.4607
			& 0.3100 & 0.8050 & 0.1159 & 0.0416 & \underline{0.4093} & 0.4817 \\
			
			DDFM
			& 0.2685 & 0.8043 & 0.3631 & 0.1543 & \underline{0.4530} & 0.7747
			& 0.3320 & 0.8054 & 0.1022 & 0.0631 & 0.3277 & 0.4224
			& 0.3202 & 0.8052 & 0.0885 & 0.0488 & 0.3488 & 0.4443 \\
			
			Text-DiFuse
			& 0.2896 & 0.8044 & 0.2109 & 0.0576 & 0.3807 & 0.3402
			& 0.2905 & 0.8044 & 0.1195 & 0.0493 & 0.3350 & 0.2589
			& 0.2929 & 0.8044 & 0.1103 & 0.0462 & 0.3381 & 0.2928 \\
			
			VDMUFusion
			& 0.2367 & 0.8039 & 0.3002 & 0.1580 & 0.4198 & 0.7105
			& 0.3381 & 0.8053 & 0.0920 & 0.0712 & 0.3191 & 0.3860
			& 0.3318 & \underline{0.8052} & 0.0791 & 0.0474 & 0.3381 & 0.3985 \\
			
			RFfusion
			& \underline{0.3021} & \underline{0.8049} & 0.2831 & 0.1217 & 0.4147 & 0.6959
			& \underline{0.3771} & \underline{0.8065} & 0.0972 & 0.0564 & 0.3855 & 0.3884
			& \underline{0.3646} & 0.8043 & 0.0836 & 0.0371 & 0.3660 & 0.3822 \\
			
			Mask-DiFuser
			& 0.2343 & 0.8039 & 0.3343 & \cellcolor{gray!20}\textbf{0.2161} & 0.4050 & 0.7271
			& 0.2572 & 0.8042 & \underline{0.2114} & 0.0860 & 0.3850 & 0.5403
			& 0.2492 & 0.8041 & \underline{0.1986} & \underline{0.0611} & 0.3686 & \underline{0.5845} \\
			
			Ours
			& \cellcolor{gray!20}\textbf{0.3505} & \cellcolor{gray!20}\textbf{0.8052} & \underline{0.4083} & \underline{0.1825} & 0.4206 & \cellcolor{gray!20}\textbf{0.7810}
			& \cellcolor{gray!20}\textbf{0.4477} & \cellcolor{gray!20}\textbf{0.8068} & \cellcolor{gray!20}\textbf{0.3698} & \cellcolor{gray!20}\textbf{0.1671} & \cellcolor{gray!20}\textbf{0.4567} & \cellcolor{gray!20}\textbf{0.7233}
			& \cellcolor{gray!20}\textbf{0.3732} & \cellcolor{gray!20}\textbf{0.8055} & \cellcolor{gray!20}\textbf{0.2199} & \cellcolor{gray!20}\textbf{0.0755} & 0.3790 & \cellcolor{gray!20}\textbf{0.6237} \\
			\hline
	\end{tabular}}
	\label{tab1}
\end{table*}

\subsection{Setup}
\textbf{Implementation Details.} The proposed model is implemented in PyTorch and optimized using the Adam optimizer. The batch size is set to 8, and the model is trained for a total of 100 epochs. The initial learning rate is fixed to 0.0001 and then decayed using a multi-step scheduler. The hyperparameter $\lambda$ is set to 10, while $\gamma$ and $\phi$ are set to 20 and 10, respectively. All experiments were implemented on a computer equipped with dual NVIDIA RTX 4090 GPUs.

\textbf{Datasets.} We evaluate the proposed method on the infrared–visible image fusion dataset M3FD \cite{liu2022target} and on PET–MRI pairs from the Harvard medical image fusion dataset. Three degradation scenarios are considered: noise, blur, and a more challenging composite degradation case. In the first two scenarios, we assume that both source images are affected by the same type of degradation. In the composite case for infrared–visible fusion, the infrared image is corrupted by noise, blur, and low resolution, while the visible image suffers only from noise and blur. For PET–MRI fusion, the PET image is subjected to all three degradations, whereas the MRI image is degraded by noise and blur only.

\textbf{Metrics and SOTA Competitors.} We adopt six widely used objective metrics for image fusion to evaluate the performance of the proposed method, including $Q_{MI}$, $Q_{NCIE}$, $Q^{AB/F}$, $Q_P$, $Q_{CB}$, and $Q_W$. In addition, we compare against eight representative fusion approaches, namely the CNN based methods IFCNN \cite{zhang2020ifcnn}, U2Fusion \cite{xu2020u2fusion}, and MURF \cite{xu2023murf}, as well as the diffusion based methods DDFM \cite{zhao2023ddfm}, Text-DiFuse \cite{zhang2024text}, VDMUFusion \cite{shi2024vdmufusion}, RFfusion \cite{wang2025efficient} and Mask-DiFuser \cite{tang2025mask}. For different degradation scenarios, we first apply existing image restoration algorithms corresponding to each type of degradation, including denoising \cite{vaksman2020lidia}, deblurring \cite{zamir2022restormer}, and super-resolution \cite{zhang2024transcending} methods, to recover the source images. The restored images are then fused using the aforementioned competing methods. For the composite degradation scenario, we sequentially apply these restoration algorithms in the above order to reconstruct the source images before fusion.

\begin{figure*}[htbp]
	\centering
	\includegraphics[width=0.9\textwidth]{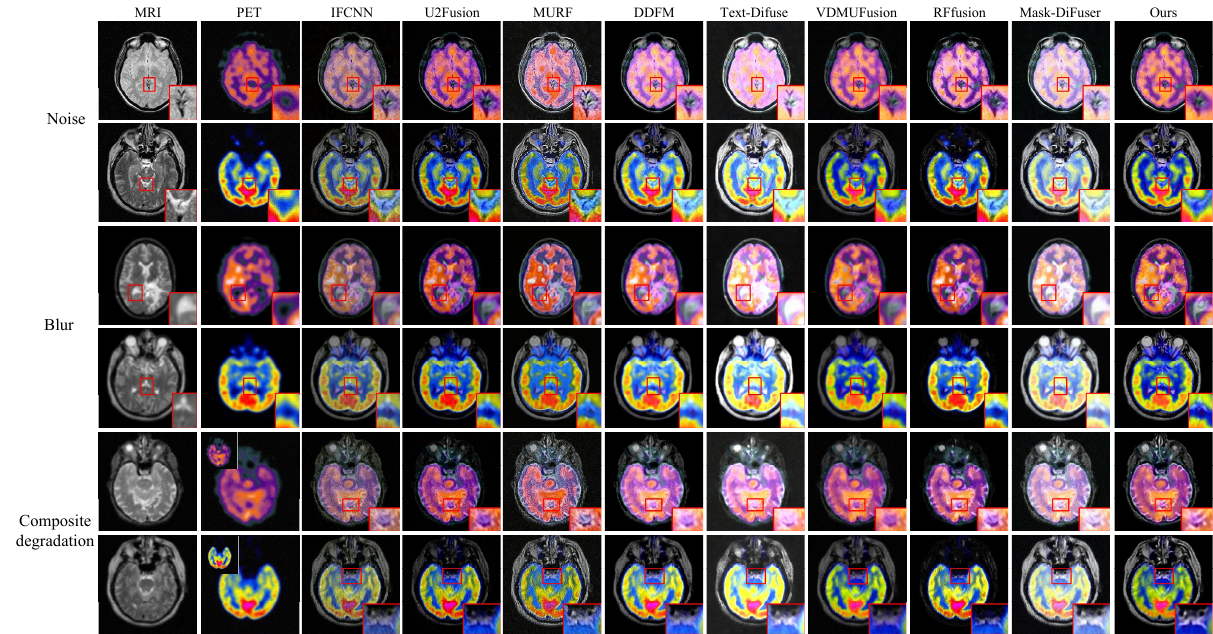} % 替换为你的图片文件名
	\caption{Qualitative comparison of different fusion methods on the Harvard dataset. For the comparison methods, we first apply the corresponding restoration algorithms (e.g., denoising, deblurring, super-resolution) to process the images, and then perform fusion. The proposed method directly reconstructs the fusion results from the degraded source images. In the third degradation scenario, the PET image is of low resolution and we have enlarged it, the original image is shown in the top-left corner.}
	\label{fig4}
\end{figure*}

\begin{table*}[t]
	\centering
	\caption{Objective fusion metrics of various methods under different degradation scenarios on the Harvard dataset (Bold and gray background: best result in each column; underline: second best).}
	\scalebox{0.56}{
		\begin{tabular}{c|cccccc|cccccc|cccccc}
			\toprule
			\multirow{2}{*}{Methods} 
			& \multicolumn{6}{c}{Noise} 
			& \multicolumn{6}{c}{Blur} 
			& \multicolumn{6}{c}{Composite degradation} \\
			\cmidrule(r){2-7} \cmidrule(r){8-13} \cmidrule(r){14-19}
			& $Q_{MI}$ & $Q_{NCIE}$ & $Q^{AB/F}$ & $Q_{P}$ & $Q_{CB}$ & $Q_{W}$
			& $Q_{MI}$ & $Q_{NCIE}$ & $Q^{AB/F}$ & $Q_{P}$ & $Q_{CB}$ & $Q_{W}$
			& $Q_{MI}$ & $Q_{NCIE}$ & $Q^{AB/F}$ & $Q_{P}$ & $Q_{CB}$ & $Q_{W}$ \\
			\midrule
			IFCNN  
			& 0.5164 & 0.8067 & 0.3902 & 0.1791 & 0.4611 & 0.7615  
			& 0.6472 & 0.8073 & 0.2692 & 0.1613 & 0.5891 & \underline{0.6752}  
			& 0.5231 & 0.8065 & 0.1853 & 0.0515 & 0.4613 & 0.5999 \\
			
			U2Fusion  
			& 0.5830 & 0.8069 & 0.3379 & 0.1658 & 0.4095 & 0.6827  
			& 0.6505 & 0.8074 & 0.2373 & 0.1523 & 0.3587 & 0.5760  
			& 0.5601 & 0.8067 & 0.1761 & 0.0628 & 0.4451 & 0.5374 \\
			
			MURF  
			& 0.4456 & 0.8061 & 0.3199 & 0.1380 & 0.3484 & 0.6431  
			& 0.6061 & 0.8072 & 0.3003 & 0.1273 & 0.3615 & 0.6936  
			& 0.4349 & 0.8059 & 0.1771 & 0.0447 & 0.3757 & 0.5462 \\
			
			DDFM  
			& 0.6025 & 0.8075 & 0.3840 & 0.1908 & \cellcolor{gray!20}\textbf{0.6078} & 0.7532  
			& \cellcolor{gray!20}\textbf{0.6653} & 0.8077 & 0.2507 & \underline{0.1690} & 0.5810 & 0.6208  
			& \cellcolor{gray!20}\textbf{0.6353} & 0.8071 & 0.1971 & 0.0715 & \cellcolor{gray!20}\textbf{0.5878} & 0.6032 \\
			
			Text-DiFuse  
			& 0.5324 & 0.8074 & 0.4026 & 0.1916 & 0.2982 & 0.6810  
			& 0.5547 & 0.8073 & 0.2839 & 0.1215 & 0.3256 & 0.5820  
			& 0.5382 & 0.8073 & 0.1810 & 0.0773 & 0.2458 & 0.4893 \\
			
			VDMUFusion  
			& 0.5737 & 0.8074 & 0.3451 & 0.1858 & 0.3617 & 0.7245  
			& \underline{0.6517} & \cellcolor{gray!20}\textbf{0.8079} & 0.2138 & 0.1450 & 0.4642 & 0.5825  
			& 0.5438 & 0.8071 & 0.1773 & 0.0635 & 0.3720 & 0.5611 \\
			
			RFfusion  
			& \underline{0.6085} & \underline{0.8076} & 0.2525 & 0.1235 & \underline{0.5113} & 0.5532  
			& 0.5831 & {0.8075} & 0.1917 & 0.1182 & \underline{0.5920} & 0.4130  
			& 0.5931 & 0.8055 & 0.1508 & 0.0577 & 0.4635 & 0.4115 \\
			
			Mask-DiFuser  
			& 0.5500 & 0.8075 & \underline{0.4210} & \underline{0.2065} & 0.3418 & \underline{0.7602}  
			& 0.5998 & {0.8076} & \underline{0.2914} & 0.1621 & 0.3029 & 0.6591  
			& 0.5151 & \underline{0.8072} & \underline{0.2114} & \underline{0.0650} & 0.3233 & \underline{0.6148} \\
			
			Ours  
			& \cellcolor{gray!20}\textbf{0.6171} & \cellcolor{gray!20}\textbf{0.8078} & \cellcolor{gray!20}\textbf{0.4258} & \cellcolor{gray!20}\textbf{0.2436} & 0.4595 & \cellcolor{gray!20}\textbf{0.7892}  
			& 0.6469 & \underline{0.8077} & \cellcolor{gray!20}\textbf{0.3855} & \cellcolor{gray!20}\textbf{0.2016} & \cellcolor{gray!20}\textbf{0.6281} & \cellcolor{gray!20}\textbf{0.7885}  
			& \underline{0.6019} & \cellcolor{gray!20}\textbf{0.8073} & \cellcolor{gray!20}\textbf{0.2956} & \cellcolor{gray!20}\textbf{0.1258} & \underline{0.4873} & \cellcolor{gray!20}\textbf{0.7344} \\
			\bottomrule
	\end{tabular}}
	\label{tab2}
\end{table*}

\subsection{Infrared and Visible Image Fusion}
\textbf{Qualitative Comparison.} The qualitative results for the infrared–visible image fusion task are presented in Fig. \ref{fig3}. For each degradation scenario, we show two representative examples and provide a zoomed-in patch in the bottom-right corner to better inspect local details. 
It can be observed that cascaded restoration and fusion pipelines fail to fully eliminate degradations, leaving residual artifacts such as blurred structures and detail loss. In contrast, our proposed method directly reconstructs high-quality fused images from degraded inputs. As shown in the visual comparisons, our approach clearly demonstrates superior preservation of color fidelity and detail information. Specifically, it effectively avoids the color distortion introduced by competing methods like Mask-DiFuser (e.g., the tree regions in the first row) and extracts more precise fine details (e.g., the text and car regions in the third and last rows). Finally, both the thermal radiation from infrared images and the rich textures from visible images are optimally preserved in our outputs.

\textbf{ Quantitative Comparison.} The corresponding quantitative results are summarized in Table \ref{tab1}. It can be observed that the proposed method achieves the best or near-best performance on most metrics across all three degradation scenarios. In particular, under the more challenging deblurring and composite degradation settings, our approach significantly outperforms existing methods in terms of $Q_{MI}$ $Q_{NCIE}$, $Q^{AB/F}$, $Q_{P}$, and $Q_{W}$. These results demonstrate that the proposed degradation-aware diffusion framework not only effectively mitigates the adverse effects of noise, blur, and resolution mismatch, but also delivers higher-quality fusion in terms of information preservation and structural detail reconstruction.

\subsection{Medical Image Fusion}

\textbf{Qualitative Comparison.} Fig.~\ref{fig4} shows the qualitative results for the proposed PET-MRI fusion method across different degradation scenarios. Although restoration algorithms reduce some degradation effects, they cannot completely eliminate issues such as blurring and detail loss. Existing fusion methods are also unable to fully optimize fusion results under these conditions. In contrast, the proposed method effectively reconstructs high-quality fused images, preserving both the PET’s metabolic information and the MRI’s structural details. The results demonstrate that the proposed approach outperforms the competing methods in terms of both information fidelity and structural clarity, even in the presence of degradation.

\begin{figure}[htbp]
	\centering
	\includegraphics[width=0.85\linewidth]{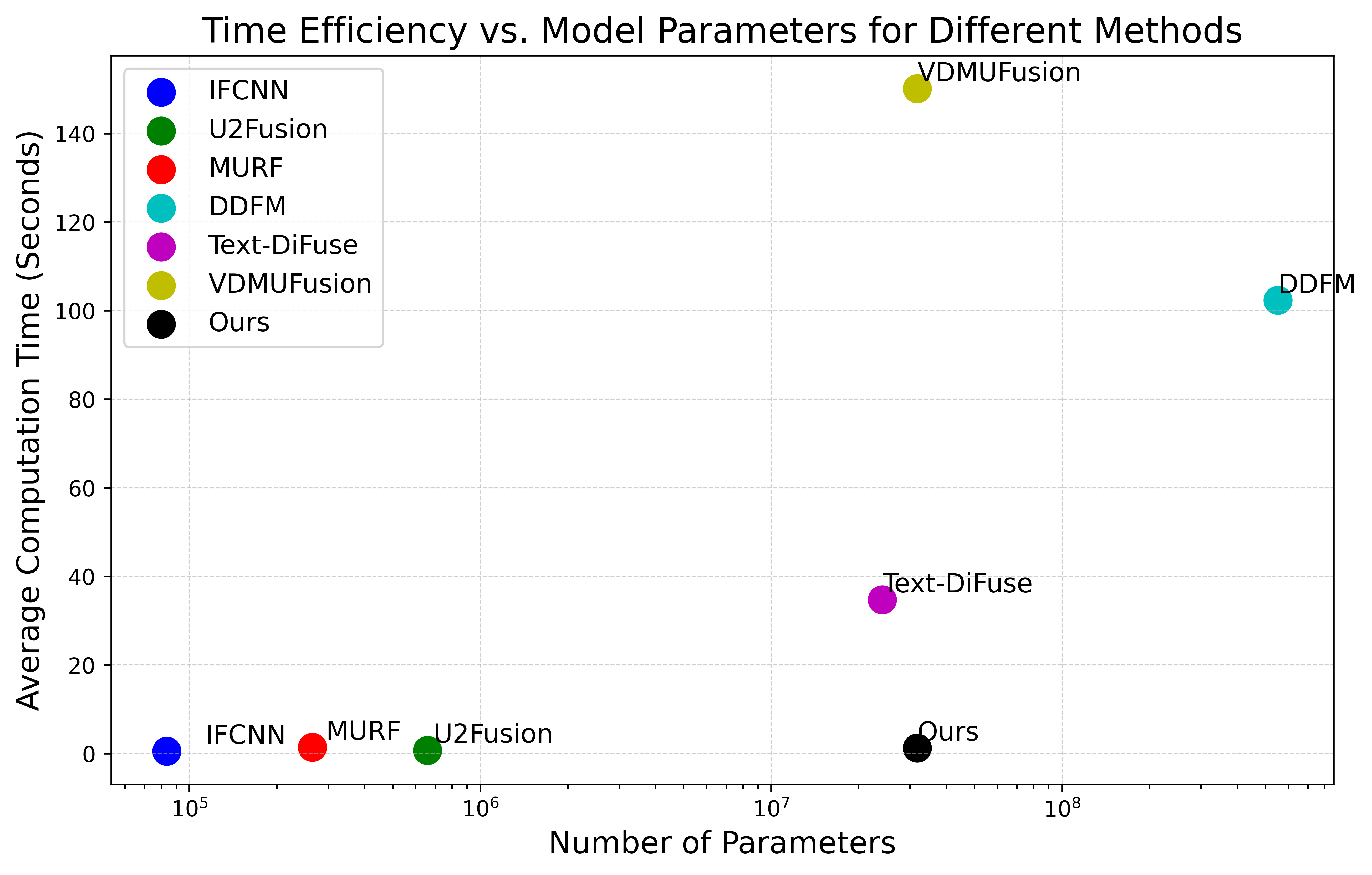} % 替换为你的图片文件名
	\caption{Comparison of time efficiency and model parameters across different fusion methods.}
	\label{fig5}
\end{figure}

\textbf{ Quantitative Comparison.} The quantitative results, summarized in Table~\ref{tab2}, demonstrate that the proposed method outperforms existing fusion approaches across all degradation scenarios. Specifically, our method achieves the highest scores in terms of $Q_{AB/F}$, $Q_{P}$, and $Q_W$ in all three degradation conditions (denoising, deblurring, and composite degradation). In particular, under the more challenging composite degradation scenario, the proposed method shows significant improvements in metrics such as $Q^{AB/F}$ and $Q_P$, highlighting its robustness in handling complex degradations. These results validate the effectiveness of the proposed degradation-aware diffusion framework in maintaining high fusion quality while preserving both structural and informational integrity.

\subsection{Efficiency and Model Parameters Analysis}

The time efficiency and model parameters analysis is shown in Fig.~\ref{fig5}. Neural network-based methods, such as IFCNN and U2Fusion, exhibit fast run times, making them highly efficient for tasks requiring speed. However, diffusion-based methods like DDFM, Text-DiFuse, and VDMUFusion generally have higher time costs due to their iterative nature, resulting in slower performance. Our proposed method strikes a balance by significantly improving time efficiency compared to traditional diffusion models. Although it cannot match the speed of neural networks, it is much faster than conventional diffusion approaches, making it a competitive choice for tasks requiring both quality and efficiency. Despite being slower than neural networks, our method offers a good trade-off between computational cost and fusion accuracy. Furthermore, regarding the number of parameters, the proposed model delivers strong performance without the need for a highly complex architecture, and it is not particularly demanding in terms of computational resources.

\subsection{Ablation Experiments}

\begin{figure}[htbp]
	\centering
	\begin{minipage}{0.48\linewidth}
		\centering
		\includegraphics[width=\textwidth]{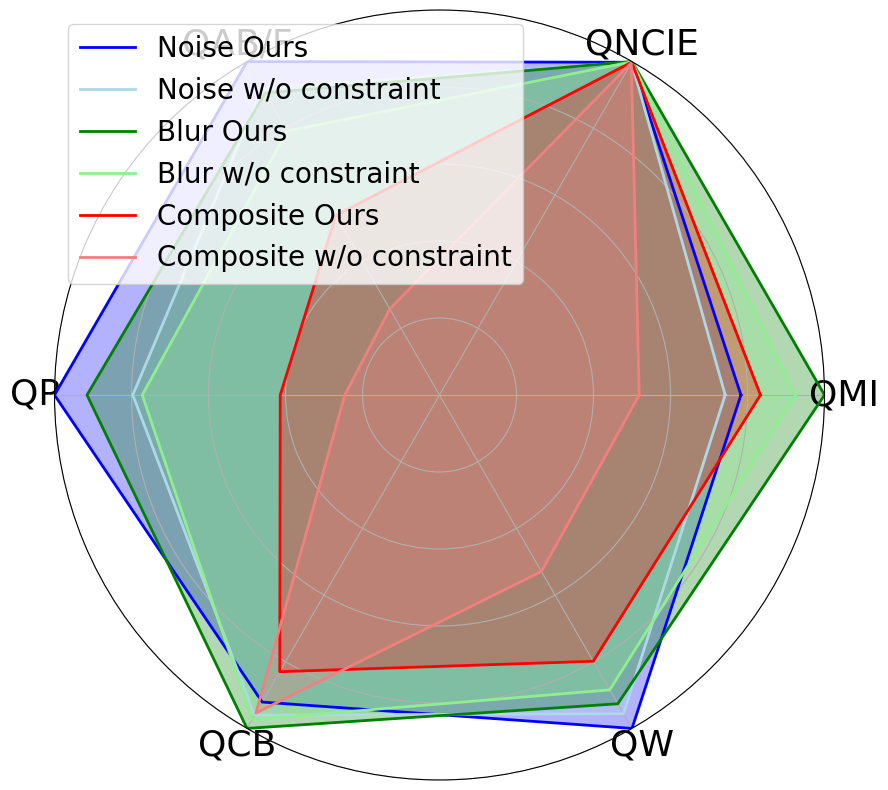}  % Replace with your image file
		\subcaption{} \label{fig:subfig1}
	\end{minipage}%
	\hspace{6pt}
	\begin{minipage}{0.48\linewidth}
		\centering
		\includegraphics[width=\textwidth]{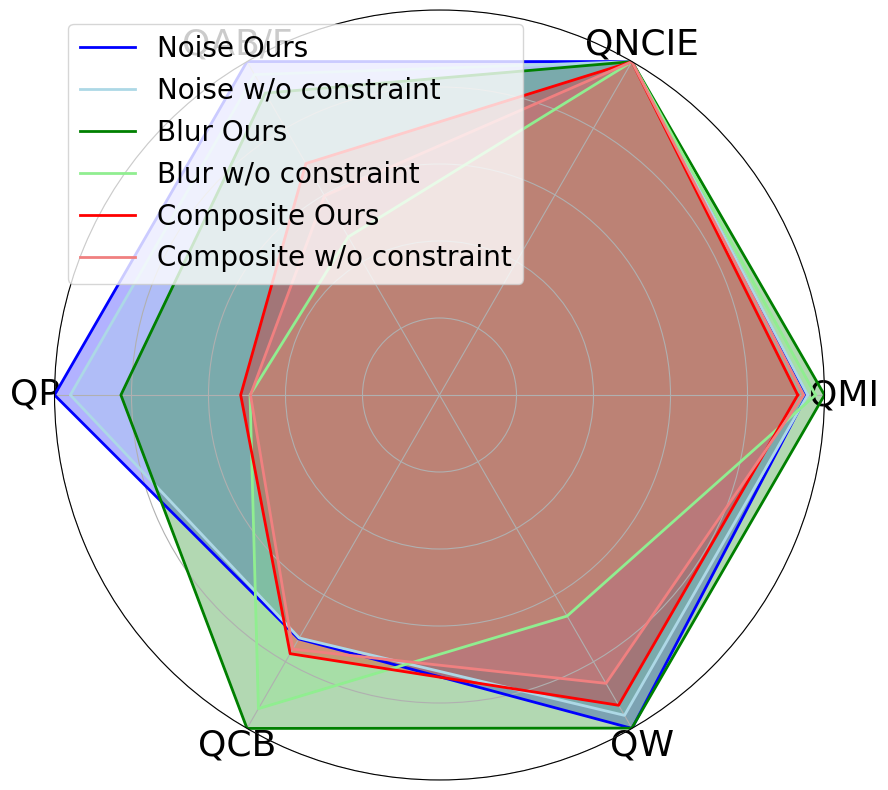}  % Replace with your image file
		\subcaption{} \label{fig:subfig2}
	\end{minipage}
	\caption{Ablation study results on two datasets. (a) represents the M3FD dataset, and (b) represents the Harvard dataset.}
	\label{fig6}
\end{figure}

In the ablation study, we compared the performance of the model with and without the joint constraint across two datasets and various degradation scenarios. The results, shown in Fig. \ref{fig6}, demonstrate that adding the joint constraint significantly improves the performance in most of the evaluation metrics, including $Q_{MI}$, $Q_{NCIE}$, $Q_{AB/F}$, $Q_P$, $Q_{CB}$, and $Q_W$. The method with the joint constraint outperforms the unconstrained model, especially in terms of reconstruction accuracy and detail preservation. This highlights the effectiveness of the joint constraint in enhancing the fusion quality and recovery of fine details, proving its importance in handling complex degradation scenarios. Details of the ablation experiments and the performance validation on downstream tasks can be found in Supplementary Material D and E.

\section{Conclusion}
In this work, we propose a degradation-aware diffusion framework for image fusion that effectively addresses various degradation scenarios, such as noise, blur, and low resolution. Extensive qualitative and quantitative evaluations demonstrate that our method outperforms existing fusion techniques, particularly in terms of preserving fine details and structural integrity in the fused images. While traditional diffusion-based methods are typically slower than neural network-based approaches, our framework achieves a notable improvement in time efficiency, approaching the speed of neural networks while maintaining superior fusion quality. This makes our approach a highly competitive solution for applications requiring both computational efficiency and high-fidelity image fusion. Furthermore, this work provides a novel perspective on diffusion-based image fusion, offering a more efficient and accessible way to leverage diffusion models for image fusion tasks.

\section*{Acknowledgement}
This work was supported in part by the National Natural Science Foundation of China under Grant 62576132 and Grant U23A20294, in part by the Fundamental and Interdisciplinary Disciplines Breakthrough Plan of the Ministry of Education of China under Grant JYB2025XDXM109, and in part by the Yunnan Fundamental Research Projects under Grants 202301AV070004 and 202501AS070123.

{
    \small
    \bibliographystyle{ieeenat_fullname}
    \bibliography{main}
}

\clearpage
\appendix
\clearpage
\setcounter{page}{1}
\maketitlesupplementary

\section{Diffusion Models}
\subsection{Denoising Diffusion Probabilistic Models}

Denoising Diffusion Probabilistic Models (DDPM) are a class of generative models that rely on a forward process of gradually adding noise to data, followed by a reverse process to recover the original data. The key idea behind DDPM is to model the distribution of the data using a Markov chain, where each step in the chain progressively adds noise until the data is destroyed into pure noise. The reverse process then attempts to reverse this noise addition, thereby generating samples from the data distribution.

\textbf{Forward Diffusion Process.} In the forward process, noise is gradually added to the data, making it more random at each step. Formally, the forward process can be described as a sequence of noisy data points $\{x_0, x_1, \dots, x_T\}$, where the data at each step $x_T$ is obtained by adding Gaussian noise to the previous step. The forward process is typically defined by a variance schedule $\beta_1, \beta_2, \dots, \beta_T$.

The forward process can be described as:
\begin{equation}
	q(x_t|x_{t - 1})=\mathcal{N}(x_t;\sqrt{1 - \beta_t}x_{t - 1},\beta_t\textbf{I}),
\end{equation}
where $\beta_t$ controls the noise added at each step, and $x_t$ is the noisy version of the data at time step $t$.

The distribution of the data at time $t$ is then:
\begin{equation}
	q(x_t | x_0) = \mathcal{N}(x_t; \sqrt{\bar{\alpha}_t} x_0, (1 - \bar{\alpha}_t) \textbf{I}),
\end{equation}
where $\alpha_t = 1 - \beta_t$ and $\bar{\alpha}_t = \prod_{i = 1}^{t} \alpha_i$.

\textbf{Reverse Process:} The reverse process involves learning a model that can reverse the noise addition, transforming pure noise back into a sample from the data distribution. This reverse process is modeled by a neural network $\epsilon_0(x_t, t)$, which predicts the noise added at each time step.

The reverse process can be written as: 
\begin{equation}
	p_\theta(x_{t-1} | x_t) = \mathcal{N}(x_{t-1}; \mu_\theta(x_t, t), \sigma_\theta(t)^2 \textbf{I}),
\end{equation}
where $\mu_\theta(x_t, t)$ and $\sigma_\theta(t)$ are the mean and variance predicted by the neural network. The model is trained to minimize the difference between the true noise and the predicted noise at each step.

\textbf{Training Objective:} The objective is to learn the reverse process by minimizing the evidence lower bound (ELBO). This is typically done by minimizing the variational bound on the negative log-likelihood, which leads to the following loss function:
\begin{equation}
	\begin{split}
		L_{\mathrm{DDPM}} &= \mathbb{E}_{q(x_0, x_T)} \Bigg[ \sum_{t=1}^{T} \mathbb{E}_{q(x_t | x_{t-1})} \Big[ \\
		&\quad \| \epsilon_\theta(x_t, t) - \epsilon^*(x_t, t) \|^2 \Big] \Bigg],
	\end{split}
\end{equation}
where $ \epsilon^*(x_t, t)$ is the true noise added at each step.

\subsection{Denoising Diffusion Implicit Models}
Denoising Diffusion Implicit Models (DDIM) are a variant of DDPM that introduce a more efficient sampling process. While DDPM generates samples by following the reverse process step by step, DDIM allows for implicit sampling, meaning that fewer steps are required to generate samples without sacrificing sample quality. DDIM achieves this by changing the reverse diffusion process, allowing for a deterministic trajectory of the reverse process.

\textbf{Reverse Process in DDIM:} The key difference in DDIM is the deterministic nature of the reverse process. Instead of adding Gaussian noise at each step, DDIM defines a reverse process with a fixed noise schedule, leading to fewer steps needed to generate high-quality samples.

The reverse process in DDIM can be defined as:
\begin{equation}
	p_\theta(x_{t-1} | x_t) = \mathcal{N}(x_{t-1}; \mu_\theta(x_t, t), \sigma_\theta(t)^2 \textbf{I}),
\end{equation}
where the mean $\mu_\theta(x_t, t)$ is determined by the model and the noise variance $\sigma_\theta(t)$ is fixed in DDIM. The reverse process is defined in such a way that the trajectory of the reverse diffusion is deterministic, leading to a more efficient sampling procedure. DDIM leverages the forward noising formula in DDPM and the reparameterization technique to transform the original $x_t$ into a deterministic mapping form of $x_0$, and ultimately derives the following iterative formula:
\begin{equation}
	\begin{split}
		x_{t-1} &= \sqrt{\alpha_{t-1}} \left( x_t - \sqrt{1 - \alpha_t} \epsilon_{\theta}^{(t)}(x_t) \right) \frac{1}{\sqrt{\alpha_t}} \\
		&\quad + \sqrt{1 - \alpha_{t-1} - \sigma_t^2} \epsilon_{\theta}^{(t)}(x_t) + \sigma_t \epsilon_t
	\end{split}
\end{equation}

This iterative formula is typically used in the context of denoising diffusion models (e.g., DDIM). It describes the process of updating the latent variable $x_{t-1}$ based on the current latent $x_t$, the deterministic mapping $x_0$, and a noise term. The terms $\alpha_t$, $\alpha_{t-1}$, and $\theta$ are parameters related to the denoising process, and the formula ensures the iterative update is consistent with the model’s reparameterization and forward noising mechanisms.

\begin{algorithm}[]
	\caption{Efficient Degradation-Aware Diffusion Framework for Image Fusion}
	\label{alg:diffusion_fusion}
	\textbf{Input}: Degraded source images ${{\bf{y}}_1}$, ${{\bf{y}}_2}$, degradation operators ${{\bf{A}}_1}$, ${{\bf{A}}_2}$, maximum iteration steps $T$. \\
	\textbf{Output}: ${\bf{X}}_f$
	\begin{algorithmic}[1]
		\STATE Construct the joint observation vector using Eq. (12).
		\STATE Initialize the input $\hat{\bf{x}}_T$ using the weighted average of the source images.
		\FOR{$t = T$ \textbf{down to} $1$}
		\STATE Predict noise $\epsilon_\theta(\hat{\bf{x}}_t, t)$ and fusion weight ${\bf{W}_1}$ using multi-task U-Net $\theta$.
		\STATE Calculate complementary fusion weight: ${\bf{W}}_2 = 1 - {\bf{W}}_1$.
		\STATE Construct the joint degradation matrix $\hat{\bf{A}}$ via {Eq. (13)}.
		\STATE Implicitly compute the pseudoinverse $\hat{\bf{A}}^{\dagger}$ via {Eq. (15)}.
		\STATE Compute the unconstrained denoised estimate $\hat{\bf{x}}_{0|t}$ via {Eq. (16)}.
		\STATE Perform joint degradation-aware correction to obtain $\overline{\bf{x}}_{0|t}$ via {Eq. (17)}.
		\STATE Update the latent state to $\hat{\bf{x}}_{t-1}$ via {Eq. (18)}.
		\ENDFOR
		\STATE Extract the final fused image component ${\bf{X}}_f$ from $\hat{\bf{x}}_0$.
		\RETURN ${\bf{X}}_f$
	\end{algorithmic}
\end{algorithm}

%------------------------------------------------------------------------
\section{Comprehensive Analysis and Refinement of Joint Constraint Correction}
In image fusion tasks, the core idea of the joint constraint correction mechanism is to introduce multiple constraints, ensuring that the fusion image not only satisfies the observation consistency of each source image but also maintains overall fusion consistency. In this mechanism, the degradation process and fusion process are coupled together, and linear constraints ensure that the relationship between each source image and the fused image is effectively constrained.

\textbf{Joint Variables and Constraint Design:} Let the joint variable be $\boldsymbol{x} = [x_1, x_2, x_f]$, where $x_1$ and $x_2$ are the two source images, and $x_f$ is the fused image. Three types of constraints are introduced to describe the relationship between the source images and the fused image:

\textbf{Data Consistency (Two-way) Constraint:}
\begin{equation}
	\mathbf{y}_1 = A_1\mathbf{x}_1 + \mathbf{n}_1,
\end{equation}
\begin{equation}
	\mathbf{y}_2 = A_2\mathbf{x}_2 + \mathbf{n}_2,
\end{equation}
where $y_1$ and $y_2$ are the observations of the two source images, $A_1$ and $A_2$ are the degradation operators, and $n_1$ and $n_2$ are the noise terms. 

\textbf{Fusion Consistency (Linear) Constraint:} 
\begin{equation}
	\mathbf{x}_f = W_1\mathbf{x}_1 + W_2\mathbf{x}_2,
\end{equation}   

where $W_1$ and $W_2$ are the linear fusion operators, which can be learned during training. The fusion operators describe the linear relationship between the source images and the fused image. 

These constraints can be combined into an overall linear equation system, using the joint degradation matrix $A$ and the joint observation $y$. Specifically, the degradation matrices 
$A_1$ and $A_2$ can be designed according to the actual task, while the fusion operators $W_1$ and $W_2$ can be learned through training, typically frozen during the internal update at each step. The joint degradation matrix can be written in the following form:
\begin{equation} 
	\left[ \begin{array}{l}
		{{\bf{y}}_1}\\
		{{\bf{y}}_2}\\
		{{\;\bf{0}}}
	\end{array} \right] = \left[ \begin{array}{l}
		\;{{\bf{A}}_1}\;\;\;\;\;\;\;\;\;0\;\;\;\;\;\;\;\;\;0\\
		\;\;0\;\;\;\;\;\;\;\;\;\;{{\bf{A}}_2}\;\;\;\;\;\;\;0\\
		- {{\bf{W}}_1}\;\; - {{\bf{W}}_2}\;\;\;\;\;{\bf{I}}
	\end{array} \right]\left[ \begin{array}{l}
		{{\bf{X}}_1}\\
		{{\bf{X}}_2}\\
		{{\bf{X}}_f}
	\end{array} \right].
\end{equation}

\textbf{Pseudo-inverse and Correction Mechanism:} For the above linear constraints, the goal of the correction process is to find a new solution $x^*$ that satisfies the constraints and minimizes the Euclidean distance to the initial solution $x_0$. Geometrically, this problem is equivalent to orthogonally projecting the initial solution $x_0$ onto the hyperplane defined by the linear equation $Ax = y$, where A is the joint matrix. The correction amount can be obtained by solving the following linear equation system:
\begin{equation}
	Ax^* = y,
\end{equation} 
Or write it in an optimized form: 
\begin{equation}
	\mathbf{x}^* = \operatorname{arg\,min}_{z} \|z - x_0|_t\|^2 \quad \text{s.t. } Az = y.
\end{equation}                                                                                     In numerical computation, we typically do not directly calculate the pseudo-inverse $A^\dagger$ because it may be expensive and unstable. Instead, we compute the correction amount:   
\begin{equation}
	\mathbf{x}^* = \mathbf{x}_0 - \Delta \mathbf{x},
\end{equation}                                                                                      where $\Delta \mathbf{x}$ is the correction amount, representing the shift from the initial solution to the optimal solution that satisfies the constraints. Specifically, the solution to the above equation is:
\begin{equation}
	x^{\star} = x_{0} - A^{\dagger}(Ax_{0} - y).
\end{equation} 
The correction amount is typically solved using Conjugate Gradient (CG), which is an efficient iterative method for large-scale problems (e.g., high-resolution images).

Joint modeling and correction offer significant benefits by ensuring consistency and coherence across multiple data sources. By simultaneously considering various constraints, this approach maintains the interdependencies between source images and the fused image, leading to more accurate and realistic results. It efficiently integrates available information, handles complex degradations, and improves computational efficiency. Moreover, joint correction minimizes errors by ensuring that the solution satisfies all constraints, even in the presence of noisy or incomplete data. This makes joint modeling particularly effective in unsupervised learning tasks and large-scale applications. However, directly using the Conjugate Gradient method for explicit computation is not feasible in practice due to the risk of memory explosion and high computational costs, as the method requires multiple iterations. Therefore, we employ an alternative approach to obtain the pseudo-inverse based on the solution of the equation. The details of this method can be found in Section 3 of the main text.                                                                                                                                                                                                                                                                                                                                                                                                                                                                                                                                                                                                                                                                                                                                                                                                                                                                                                                                                                                                                                                                                                                                                                                                                                                                                                                                                                                                                                                                 

\section{Degradation Definition}
Here’s a detailed description of the three operations—noise addition, blurring, and super-resolution—and their corresponding $A$ (degradation matrices) and $A^\dagger$ (pseudo-inverse) in the context of the proposed model.

\textbf{Noise Addition:} Noise addition is a common degradation process where random noise is introduced to the original image. Mathematically, this operation can be expressed as:
\begin{equation}
	y = Ax + n.
\end{equation} 
The pseudo-inverse of the degradation matrix $A$ in the case of noise addition is simply the identity matrix. In the diffusion process, to reduce the impact of noise, we apply an additional intensity control coefficient to the noise-containing correction term. This approach is inspired by the DDNM.

\textbf{Blurring:} In image restoration tasks, the blur operator $A$ typically represents a linear degradation process, where an image $x$ is convolved with a blurring kernel $k$ to produce a degraded image $y$, i.e., $y = Ax = k*x$, where * denotes the convolution operation. This degradation process can be seen as a linear system where the blurring kernel $k$ acts as a filter that removes or distorts certain image details.

\begin{figure*}[htbp]
	\centering
	\includegraphics[width=0.8\textwidth]{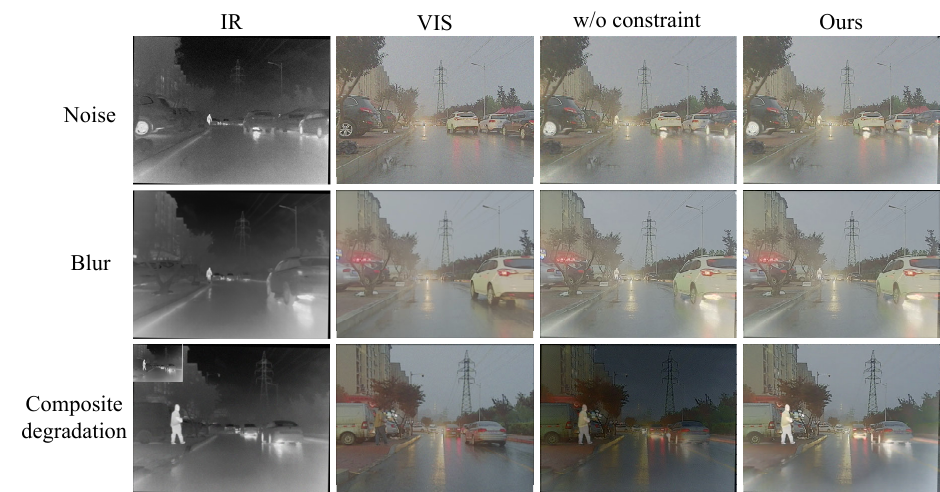} % 替换为你的图片文件名
	\caption{Results with and without the joint constraint correction mechanism under different degradation scenarios on M3FD dataset.}
	\label{M3FD_ablation}
\end{figure*}

To restore the original image $x$ from the blurred observation $y$, we need to compute the pseudo-inverse $Ap$ of the blur operator. In the frequency domain, Wiener convolution provides an optimal solution to this problem by minimizing the mean square error between the true and estimated images. The Wiener filter in the frequency domain is given by:
\begin{equation}
	H(\omega)=\frac{H^*(\omega)}{|H(\omega)|^2+\gamma},
\end{equation} 
Where $H(\omega)$ is the Fourier transform of the blurring kernel $k$, $\gamma$ is a regularization term that accounts for noise in the observation, and $H^*(\omega)$ is the complex conjugate of $H(\omega)$. This filter effectively acts as a frequency-domain approximation of the pseudo-inverse, restoring the image by compensating for the blurring and noise.

The Wiener convolution approach is ideal for this problem for several reasons:
1) Linear Degradation Model: The blur operator is linear, meaning the relationship between the observed and true image can be captured using a linear filter, making Wiener convolution a suitable choice for solving the inverse problem.
2) Frequency Domain Efficiency: By working in the frequency domain, the Wiener filter takes advantage of the fast Fourier transform (FFT), significantly speeding up the computation of the pseudo-inverse.
3) Noise Suppression: The regularization term $\gamma$ in the Wiener filter helps mitigate the amplification of noise, ensuring that the restored image is not overly influenced by noise in the observation. Thus, using Wiener convolution to compute $Ap$ provides an effective and computationally efficient method to reverse the blurring process and recover the original image, especially in the presence of noise.

\begin{figure*}[htbp]
	\centering
	\includegraphics[width=0.78\textwidth]{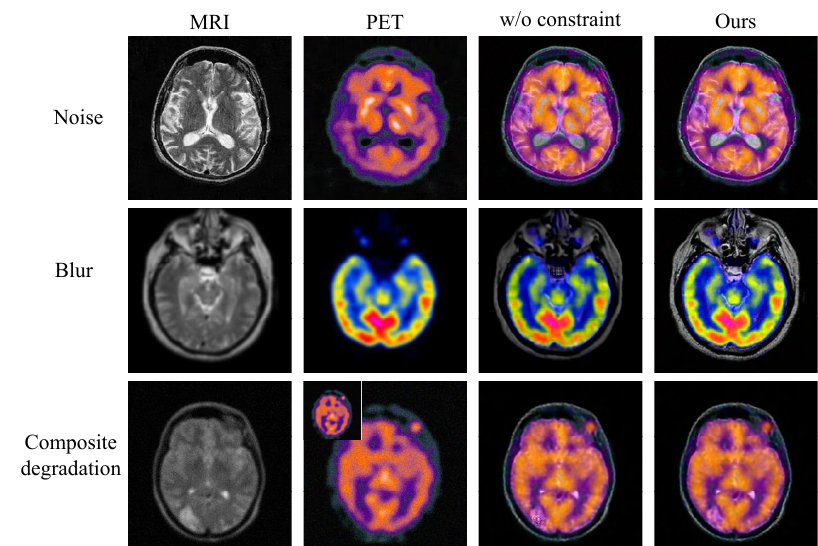} % 替换为你的图片文件名
	\caption{Results with and without the joint constraint correction mechanism under different degradation scenarios on PET-MRI dataset.}
	\label{PET_MRI_ablation}
\end{figure*}

\textbf{Low Resolution:} The degradation operator $A$ represents the downsampling operation applied to the original image to simulate a lower-resolution observation. In this case, the degradation process is modeled by an adaptive average pooling operation, which reduces the image resolution by a factor determined by the scaling parameter. The pseudo-inverse operator $Ap$ corresponds to the upsampling operation, where the low-resolution image is transformed back to a higher-resolution image. This operation is achieved by the PatchUpsample function, which increases the spatial dimensions of the image. The PatchUpsample function upsamples the low-resolution image $x$ by a factor of scale. This operator restores the image to its higher resolution by expanding the spatial dimensions (height and width) of the input image, effectively reversing the downsampling process. It does this by distributing the pixel values of the low-resolution image into the larger output grid.

\renewcommand{\arraystretch}{1.5}
\begin{table*}[htbp]
	\centering
	\setlength{\tabcolsep}{2pt} 
	\caption{Ablation results for M3FD and PET-MRI datasets under different degradation scenarios. The best values for each metric are highlighted in light gray.}
	\scalebox{0.8}{
		\begin{tabular}{c|ccccccc|cccccc}  % Ensure all vertical lines are defined
			\hline
			& \multicolumn{7}{c|}{M3FD} & \multicolumn{6}{c}{PET-MRI} \\  % Added | for vertical lines at the ends
			\hline
			& \multicolumn{2}{c}{Noise} & & \multicolumn{2}{c}{Blur} & \multicolumn{2}{c|}{Composite degradation} & \multicolumn{2}{c}{Noise} & \multicolumn{2}{c}{Blur} & \multicolumn{2}{c}{Composite degradation} \\
			\hline
			Metrics & Ours & w/o constraint & & Ours & w/o constraint & Ours & w/o constraint & Ours & w/o constraint & Ours & w/o constraint & Ours & w/o constraint \\
			\hline
			$Q_{MI}$ & \cellcolor{gray!20} \textbf{0.3505} & 0.3322 & & \cellcolor{gray!20} \textbf{0.4477} & 0.4140 & \cellcolor{gray!20} \textbf{0.3732} & 0.2322 & 0.6171 & \cellcolor{gray!20} \textbf{0.6185} & \cellcolor{gray!20} \textbf{0.6469} & 0.6318 & 0.6019 & \cellcolor{gray!20} \textbf{0.6116} \\
			$Q_{NCIE}$ & \cellcolor{gray!20} \textbf{0.8052} & 0.8050 & & \cellcolor{gray!20} \textbf{0.8068} & 0.8062 & \cellcolor{gray!20} \textbf{0.8055} & 0.8038 & \cellcolor{gray!20} \textbf{0.8078} & 0.8077 & \cellcolor{gray!20} \textbf{0.8077} & 0.8071 & 0.8073 & \cellcolor{gray!20} \textbf{0.8074} \\
			$Q^{AB/F}$ & \cellcolor{gray!20} \textbf{0.4083} & 0.3759 & & \cellcolor{gray!20} \textbf{0.3698} & 0.3233 & \cellcolor{gray!20} \textbf{0.2199} & 0.1054 & \cellcolor{gray!20} \textbf{0.4258} & 0.4092 & \cellcolor{gray!20} \textbf{0.3855} & 0.2019 & \cellcolor{gray!20} \textbf{0.2956} & 0.2544 \\
			$Q_P$ & \cellcolor{gray!20} \textbf{0.1825} & 0.1454 & & \cellcolor{gray!20} \textbf{0.1671} & 0.1409 & \cellcolor{gray!20} \textbf{0.0755} & 0.0449 & \cellcolor{gray!20} \textbf{0.2436} & 0.2336 & \cellcolor{gray!20} \textbf{0.2016} & 0.1199 & \cellcolor{gray!20} \textbf{0.1258} & 0.1196 \\
			$Q_{CB}$ & 0.4206 & \cellcolor{gray!20} \textbf{0.4386} & & \cellcolor{gray!20} \textbf{0.4567} & 0.4445 & 0.3790 & \cellcolor{gray!20} \textbf{0.4356} & \cellcolor{gray!20} \textbf{0.4595} & 0.4577 & \cellcolor{gray!20} \textbf{0.6281} & 0.5909 & \cellcolor{gray!20} \textbf{0.4873} & 0.4788 \\
			$Q_W$ & \cellcolor{gray!20} \textbf{0.7810} & 0.7465 & & \cellcolor{gray!20} \textbf{0.7233} & 0.6907 & \cellcolor{gray!20} \textbf{0.6237} & 0.4135 & \cellcolor{gray!20} \textbf{0.7892} & 0.7583 & \cellcolor{gray!20} \textbf{0.7885} & 0.5232 & \cellcolor{gray!20} \textbf{0.7344} & 0.6825 \\
			\hline
	\end{tabular}}%
	\label{ablation}% 
\end{table*}

\begin{figure*}[htbp]
	\centering
	\includegraphics[width=0.96\textwidth]{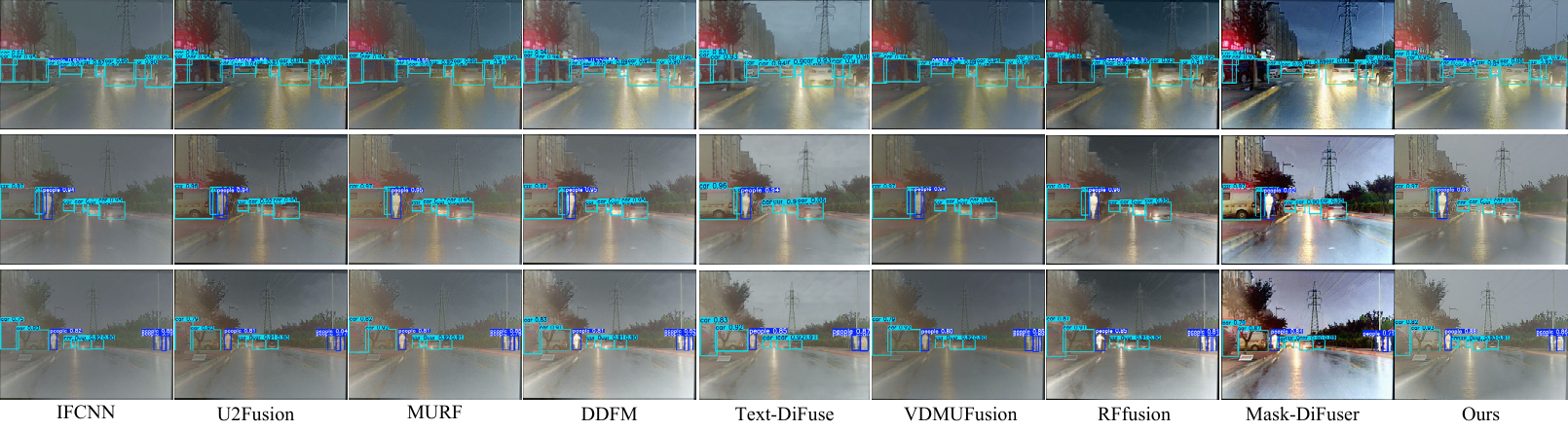} % 替换为你的图片文件名
	\caption{Qualitative detection results based on fusion images generated by different fusion methods.}
	\label{object_det}
\end{figure*}

\section{Ablation Experiments}

In the main body of the paper, we did not present the specific results and detailed metrics of the ablation experiments. These results will be provided in this supplementary material. The analysis will focus on the performance of the fusion results in three different tasks: denoising, deblurring, and compound degradation. The results on the M3FD and Harvard Medical datasets are shown in Fig~\ref{M3FD_ablation} and Fig~\ref{PET_MRI_ablation}, respectively.

\newcommand{\best}[1]{\cellcolor{gray!20}\textbf{#1}}
\begin{table}[t]
	\centering
	\caption{Detection performance comparison on M3FD dataset.}
	\resizebox{0.9\columnwidth}{!}{
		\begin{tabular}{cccc}
			\hline
			Method       & Precision & Recall  & mAP@0.5 \\
			\hline
			IFCNN        & 0.9396    & 0.7933  & 0.8906  \\
			U2Fusion     & 0.9273    & 0.7605  & 0.8700  \\
			MURF         & 0.9510    & 0.7942  & 0.8865  \\
			DDFM         & 0.9620    & 0.7062  & 0.8509  \\
			Text-Difuse  & 0.9723    & 0.6546  & 0.8172  \\
			VDMUFusion   & 0.9690    & 0.7042  & 0.8499  \\
			RFfusion     & 0.9542    & 0.7504  & 0.8705  \\
			Mask-DiFuser & 0.8769    & 0.6449  & 0.7891  \\
			Ours         & \best{0.9750} & \best{0.8005} & \best{0.9108} \\
			\hline
	\end{tabular}}
	\label{det_comparison}
\end{table}

From these figures, it is evident that removing the proposed joint constraint correction mechanism leads to a noticeable degradation in the fusion results. Specifically, the images become noisier, with less distinct details, and more edge artifacts appear. This effect is particularly pronounced in the M3FD dataset, which aligns with the intended degradation scenarios. The results suggest that, in more complex degradation conditions, the proposed correction mechanism plays a crucial role in improving the final fusion accuracy. In the denoising task, for example, the absence of the joint constraints causes the fused image to retain more noise, leading to poor preservation of structural details. Similarly, in the deblurring task, without the proposed mechanism, the image sharpness significantly decreases, and blurring artifacts become more prominent. This trend is consistent across both datasets, indicating that the joint constraint correction mechanism is particularly effective in handling more complex degradation scenarios, where traditional fusion methods struggle to provide accurate reconstructions. Especially in the compound degradation scenarios of the M3FD dataset, it is evident that the more complex the scene, the more significant the improvement in fusion accuracy brought by the constraint correction mechanism.

The objective metrics, summarized in Table~\ref{ablation}, present the results of the ablation study on the M3FD and PET-MRI datasets under different degradation scenarios (Noise, Blur, and Composite). The results show that our method outperforms the baseline (w/o constraint) across most metrics. Specifically, the proposed method achieves higher values for key metrics such as $Q_{MI}$, $Q_{NCIE}$, and $Q_P$, indicating better preservation of image details, structural integrity, and perceptual quality. These improvements are especially noticeable in more complex scenarios, such as composite degradation, where our method effectively preserves higher image quality and reduces degradation artifacts. Overall, the results highlight the significant benefits of the joint constraint correction mechanism in enhancing fusion performance across various degradation conditions.

\section{Performance on High-level Vision Task}
Image fusion is an effective form of image enhancement, whose ultimate goal is to facilitate subsequent high-level vision tasks such as object detection in video surveillance and lesion segmentation in clinical diagnosis. Better fusion quality should naturally translate into better performance on downstream vision tasks. To verify the practical utility of the proposed model, we evaluate its detection performance under the most adverse degradation scenarios by conducting object detection experiments on the M3FD dataset using detection results obtained from different fusion methods.

As reported in Table~\ref{det_comparison}, different fusion methods lead to clearly different detection performance on the M3FD dataset. Our method achieves the best results on all three metrics, with a precision of 0.9750, recall of 0.8005, and mAP@0.5 of 0.9108. Compared with the strongest baseline IFCNN, our approach improves mAP@0.5 by about 2.0 percentage points (from 0.8906 to 0.9108) while slightly increasing both precision and recall. Several competing methods, such as DDFM, Text-DiFuse and VDMUFusion, obtain relatively high precision (around 0.96–0.97) but suffer from noticeably lower recall (below 0.71) and mAP@0.5 (below 0.86), indicating that they tend to miss more targets. In contrast, our fusion model provides a more favorable balance between precision and recall, leading to the overall highest detection accuracy and confirming its effectiveness for downstream high-level vision tasks.

The visual results of the detection task are shown in Fig.~\ref{object_det}. For the comparison methods such as IFCNN and U2Fusion, the pedestrian regions in the fused images are relatively dark, leading to suboptimal detection performance, while other methods also suffer from noticeable distortions in different areas, indicating limited fusion accuracy. In contrast, the results of our method are superior to all competitors both in terms of degradation removal and fusion quality. Most detected regions in our fused images achieve higher confidence scores than those obtained by all comparison methods, which further demonstrates the potential of the proposed approach for practical applications.

\section{Analysis of the Parameter $T$}
\begin{table}[t]
	\centering
	\caption{Quantitative comparison of different $T$ values.}
	\label{tab:ablation_t}
	\resizebox{0.98\linewidth}{!}{
		\begin{tabular}{c|ccccc}
			\hline
			Metrics & $T=1$ & $T=2$ & $T=3$ (Ours) & $T=4$ & $T=5$ \\
			\hline
			$Q_{MI}$   & 0.3577 & \underline{0.3721} & \textbf{0.3732} & 0.3717 & 0.3663 \\
			$Q_{NCIE}$ & 0.8053 & \underline{0.8055} & \underline{0.8055} & 0.8047 & \textbf{0.8065} \\
			$Q^{AB/F}$ & 0.2061 & 0.2184 & \textbf{0.2199} & \underline{0.2187} & 0.2185 \\
			$Q_{P}$    & 0.0618 & 0.0660 & \textbf{0.0755} & \underline{0.0693} & 0.0689 \\
			$Q_{CB}$   & 0.3216 & 0.3587 & \textbf{0.3790} & \underline{0.3768} & 0.3673 \\
			$Q_{W}$    & 0.6192 & \underline{0.6221} & \textbf{0.6237} & 0.6178 & 0.6133 \\
			Runtime (s)& \textbf{0.1425} & 0.2201 & 0.3024 & 0.3768 & 0.4589 \\
			\hline
	\end{tabular}}
\end{table}

To investigate the impact of the iteration number $T$ on the final fusion performance, we conduct an ablation study with $T$ ranging from 1 to 5, as reported in Table \ref{tab:ablation_t}. It can be observed that when $T$ increases from 1 to 3, the objective metrics exhibit a continuous upward trend, indicating that the iterative mechanism effectively refines and enhances the fusion quality. The performance reaches its peak at $T=3$, where our method achieves the best results on most metrics, including $Q_{MI}$, $Q^{AB/F}$, $Q_{P}$, $Q_{CB}$, and $Q_{W}$. However, further increasing the iteration steps (e.g., $T=4$ and $T=5$) does not bring additional performance gains and even leads to slight metric degradation. This phenomenon may be attributed to potential over-smoothing or accumulated errors during the prolonged iterative process. Furthermore, the inference runtime increases linearly with $T$. Therefore, taking both fusion quality and computational efficiency into consideration, we set $T=3$ as the default configuration for our model.

\end{document}